\pdfoutput=1

\documentclass[11pt]{article}

\usepackage[final]{acl} %

\usepackage{times}
\usepackage{latexsym}
\usepackage{booktabs}
\usepackage{amssymb}
\usepackage{algorithm}
\usepackage{algorithmic}
\usepackage{bbm}
\usepackage{bm}
\usepackage{amsmath}
\usepackage{graphicx}

\usepackage{amsfonts}
\usepackage{textcomp}
\usepackage{xcolor}
\usepackage{colortbl}

\definecolor{Gray}{gray}{0.9}

\usepackage{caption}
\usepackage{multirow}
\usepackage{float}
\usepackage{subcaption}
\usepackage{arydshln}

\usepackage[T1]{fontenc}

\usepackage[utf8]{inputenc}

\usepackage{microtype}

\usepackage{inconsolata}

\title{GLTW: Joint Improved \underline{G}raph-Transformer Encoder and \underline{L}LM via \underline{T}hree-\underline{W}ord Language for Knowledge Graph Completion}
\author{
\textbf{Kangyang Luo$^{\spadesuit}$, Yuzhuo Bai$^{\spadesuit}$, Cheng Gao$^\spadesuit$, Shuzheng Si$^{\spadesuit}$, Yingli Shen$^\spadesuit$}\\
\textbf{Zhu Liu$^{\spadesuit}$, Zhitong Wang$^{\spadesuit}$, Cunliang Kong$^\spadesuit$, Wenhao Li$^\spadesuit$, Yufei Huang$^\spadesuit$} \\ 
\textbf{Ye Tian$^\diamondsuit$, Xuantang Xiong$^\diamondsuit$, Lei Han$^\diamondsuit$, Maosong Sun\thanks{Corresponding author}$^{\spadesuit}$$^\clubsuit$$^\bigstar$ }\\
$^{\spadesuit}$Department of Computer Science and Technology, Tsinghua University 
\\ $^\clubsuit$Institute for AI, Tsinghua University \quad $^\diamondsuit$Tencent Robotics X \\ $^\bigstar$Jiangsu Collaborative Innovation Center for Language Ability
}

\begin{document}
\maketitle
\begin{abstract}
Knowledge Graph Completion (KGC), which aims to infer missing or incomplete facts, is a crucial task for KGs. 
However, integrating the vital structural information of KGs into large language models (LLMs) and outputting predictions deterministically remains challenging. 
To address this, we propose a new method called \textbf{GLTW}, which encodes the structural information of KGs and merges it with LLMs to enhance KGC performance. 
Specifically, we introduce an improved Graph Transformer (\textbf{iGT}) that effectively encodes subgraphs with both local and global structural information and inherits the characteristics of language model, bypassing training from scratch. 
Also, we develop a subgraph-based multi-classification training objective, using all entities within KG as classification objects, to boost learning efficiency.
Importantly, we combine iGT with an LLM that takes KG language prompts as input.
Our extensive experiments on various KG datasets show that GLTW achieves significant performance gains compared to SOTA baselines. 

\end{abstract}

\section{Introduction}
Knowledge graphs (KGs) are pivotal resource for a multitude of knowledge-intensive intelligent tasks (e.g., question 
answering~\cite{zhai2024towards}, recommendation systems~\cite{zhao2024breaking}, planning~\cite{wang2024learning}, and reasoning~\cite{chen2024plan}, among others). They are composed of a vast number of triplets in the format of $(h, r, t)$, where $h$ and $t$ represent the head and tail entities, respectively, and $r$ denotes the relationship connecting these two entities.
However, popular existing KGs, such as Freebase~\cite{bollacker2008freebase}, WordNet~\cite{miller1995wordnet}, and WikiData~\cite{vrandevcic2014wikidata}, suffer from a significant drawback: the presence of numerous incomplete or missing triplets, thereby giving rise to the task of KG Completion (KGC). 
KGC aims to accurately predict the missing triplets by leveraging known entities and relations for effectively enhancing KGs.

In recent years, with super-sized training corpora and computational cluster resources, Large Language Models~(LLMs) have developed rapidly and enabled state-of-the-art performance in a wide range of natural language tasks~\cite{touvron2023llama, qin2023chatgpt,liu2024deepseek}.
Consequently, certain studies have applied LLMs to KGC tasks. For instance, \cite{yao2023exploring, zhu2024llms, wei2024kicgpt} utilize zero/few-shot in-context learning~(ICL) to accomplish KGC, while \cite{li2024contextualization, xu2024multi} leverage LLMs to enhance the descriptions of entities and relations in KGs, thereby improving text-based KGC methods~\cite{yao2019kg, zhang2020pretrain, wang2022simkgc, liu2022know, wang2022language, yang2024knowledge}.  
Intuitively, 
integrating non-textual structured information appropriately can augment LLMs' understanding and representation of KGs.
For example,  \cite{zhang2024making, liu2024can, guo2024mkgl} combine graph-structured information with LLMs to boost KGC tasks. 


Yet, they either use traditional embedding-based KGC methods~\cite{bordes2013translating, lin2015learning, sun2019rotate, balavzevic2019tucker} that only consider internal links of triplets or rely on Graph Neural Networks~(GNNs)~\cite{bronstein2021geometric, corso2020principal} that merely encode local subgraphs, thus missing out on global structural knowledge. 
Also, LLMs, typically used for generative tasks, have long been troubled by hallucination~\cite{ji2023survey, rawte2023survey}.
In contrast, the prediction targets of KGC are generally confined to the given KG, making it unwise to directly integrate LLMs into KGC tasks\footnote{Notably, see Appendix~\ref{Related_work:} for more related works.}. 
In short, how to encode both local and global structural information of KGs and combine it with knowledge-rich LLMs to achieve deterministic KGC remains underexplored. 


To this end, we propose a novel method (named \textbf{GLTW}), which effectively encodes KG subgraphs with both local and global structural information and integrates LLMs in a deterministic fashion to improve the performance of KGC. 
Concretely, 
we first treat entities and relations within KG as inseparable units, adding them as tokens to the original Tokenizer, while referring to triplets as three-word sentences~\cite{guo2024mkgl}.
Subsequently, for each target triple, we extract a subgraph that encompasses both local and global structural information from the given training KG data (Section~\ref{sub_extr:sec}). 
To effectively process the subgraph, we introduce an improved Graph Transformer (\textbf{iGT}), which takes the entity and relation embeddings (initialized by a pooling operation), the relative distance matrix, and the relative distinction matrix of the subgraph as inputs, and encodes them using the enhanced graph attention mechanism (Section~\ref{iGTE_sec:}).
Furthermore, we construct multiple positive and negative triplet samples from the subgraph, which are used to build the subgraph-based multi-classification training objective with all entities within the KG as classification objects 
(Section~\ref{sub_training_obj:sec}).
Finally, we merge iGT with LLM that takes KG language prompt as input (Section~\ref{Join_igt_llm:sec}). 
To sum up, we highlight our contributions as follows:
\begin{itemize}
    \item  We formulate a novel method, GLTW, which aims to encode both local and global structural information of KG and amalgamate it with LLMs to enhance KGC performance. Note that we consider KGC as a subgraph-based multi-classification task, outputting prediction probabilities for all entities from KG at once.
    \item We introduce iGT, which simplifies the complexity of positional encoding for subgraphs, enlarges the size of subgraphs, and treats entities and relations in a differentiated yet fair manner. Importantly, it inherits the characteristics of language model, thereby avoiding training from scratch. 
    \item We conduct extensive experiments on three commonly used KG datasets~(i.e., WN18RR, FB15k-237, and Wikidata5M) to show that GLTW is highly competitive compared with other state-of-the-art baselines. Meanwhile, ablation studies demonstrate the efficacy and indispensability for core modules and key parameters.
\end{itemize}


\begin{figure*}[t]
  \centering
  \includegraphics[scale=0.8]{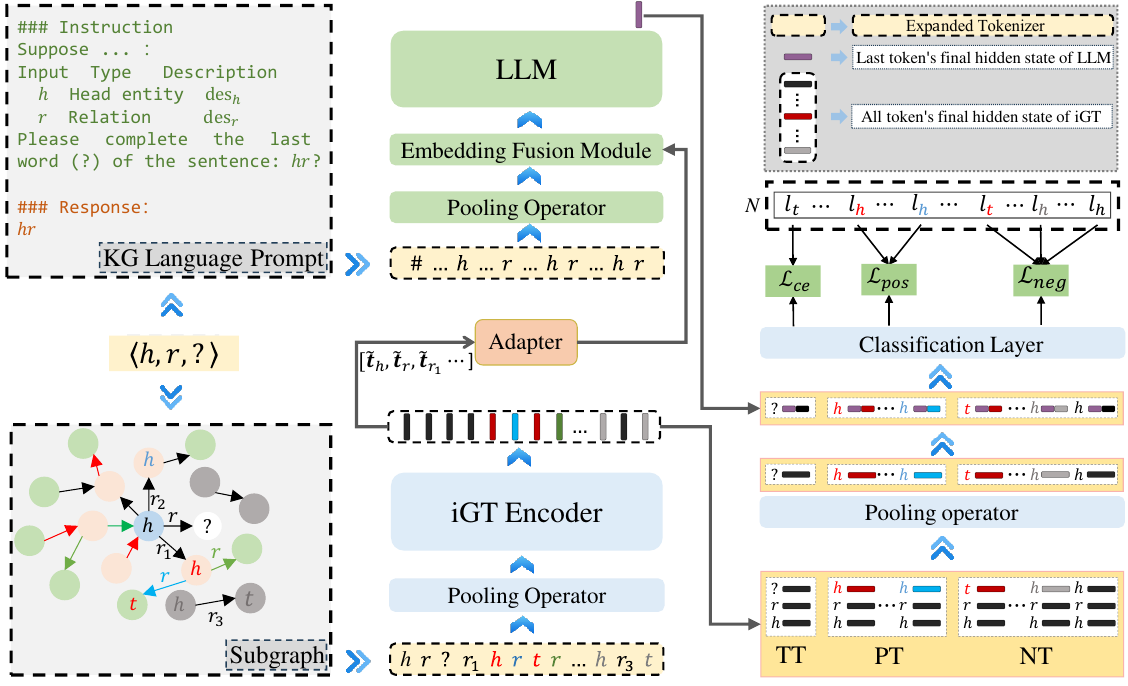}
  \caption{The pipeline of GLTW. $\mathcal{L}_{ce}$, $\mathcal{L}_{pos}$, and $\mathcal{L}_{neg}$ are loss objectives for Target Triplet (TT), Positive Triplets (PT) and Negative Triplets (NT), respectively. Notably, the $r$, $r_1$, $r_2$, and $r_3$ highlighted in black pertain to the same relation but exist in different triplets. For simplicity, ${\color{blue} h}$ and ${\color{red} h}$ can be either head or tail entities, as they are shared by multiple triplets.}
  \label{fig1:framework}
  \vspace*{-3ex}
\end{figure*}

\section{Preliminaries}
\subsection{Task Definition}
Knowledge graphs (KGs) are directed graphs that can be formally represented as $\mathcal{G}=\{\mathcal{E},\mathcal{R}, \mathcal{T}\}$, where $\mathcal{E}$ and $\mathcal{R}$ denote respectively the sets of entities and relations, and $\mathcal{T}=\{(h, r, t)\}\in \mathcal{E} \times \mathcal{R} \times  \mathcal{E}$ defines a collection of triples.
The goal of KGC is to accurately predict the incomplete triples that exist within $\mathcal{G}$. In this paper, we focus on the \textbf{link prediction} task, a key component of KGC. This task is designed to predict the missing entity $?$ in a given triple $(h,r,?)$ or $(?,r,t)$. 
We unify the link prediction task into tail entity prediction by constructing inverse relation $r^{-1} \in \mathcal{R}^{-1}$, i.e., $(t, r^{-1}, ?)$.

\subsection{Graph Transformer}
\label{GLM_sec:}
The attention mechanism~\cite{shehzad2024graph} in a graph transformer can be expressed as follows:
\begin{align}
\label{eq:at_me}
    {\rm softmax}\left(\frac{QK^\top}{\sqrt{d}}+B_P+M\right)V,
\end{align}
where $Q$, $K$, and $V$ denote the query, key and value matrices, and $d$ represents the query and key dimension. The matrices $B_P$ and $M$ serve the purposes of Positional Encoding (PE) and masking. In GLM~\cite{plenz-frank-2024-graph}, $B_P=f(P)$, where $P$ is the relative distance matrix based on Levi graph of subgraph~(as shown in Fig.~\ref{fig:p_D_gGLM}(a)-(b) in Appendix~\ref{app:P_D_gglm}), and $f$ is an element-wise function; $M$ is a zero matrix\footnote{In this paper, we focus on the global GLM ($g$GLM), which invokes an additional G2G relative position to access distant triplets and sets $M$ is a zero matrix.}. This non-invasive modification avoids pre-training from scratch and preserves compatibility with the language model parameters.

 

\subsection{Three-word Language} 
The concept of the three-word language originates from the MKGL method proposed by~\cite{guo2024mkgl}, which considers individual entities and relations as indivisible tokens and incorporates them into the LLM tokenizer (i.e., expanded tokenizer). 
For example, entity \textit{black poodle} and relation \textit{is a} are encoded as tokens \textit{<kgl: black poodle>} and \textit{<kgl: is a>}, respectively, and are employed to construct corresponding KG language prompt (see Appendix~\ref{app:Three_word_prompt}).
To prevent training these new tokens from scratch, MKGL utilizes a GNN encoder to derive their embeddings from the original tokenizer based on the textual and structural information of the entities/relations. This enables LLMs to effectively navigate and master the three-word language.



\section{Method}
In this section, we elaborate on our proposed method, GLTW, in four parts: \textit{Subgraph Extraction}, \textit{Improved Graph Transformer}, \textit{Subgraph-based Training Objective}, and \textit{Joint iGT and LLM}.
Figure~\ref{fig1:framework} illustrates the pipeline of GLTW. Notably, the KG language prompt in this paper directly follows that of MKGL~\cite{guo2024mkgl}.


\subsection{Subgraph Extraction}
\label{sub_extr:sec}
Before training or prediction, we extract a subgraph $\mathcal{G}_{sub}(h,r,t)$ for each target triplet $(h,r,t)$ from $\mathcal{G}$.  
For consistent training and prediction, we require that the subgraph only comprises triplets sampled from given $h$ and $r$, represented as $\mathcal{G}_{sub}(h,r,?)$.
$\mathcal{G}_{sub}(h,r,?)$ contains three types of triplet subsets: $T_{hr}$, $T_{h}$ and $T_{r}$, where  $T_{hr}$ and $T_{h}$ hold neighboring triplets around $(h,r,?)$, and $T_{r}$ samples distant (global) triplets with $r$. 
For $T_{hr}$ and $T_{h}$, we set the sampling radius as $l$, then $T_{hr/h}=\cup_{i=1}^l T_{hr/h}^i$.
Specifically, when $l=1$, $T_{hr}^1=\{(h, r, t^1)| t^1 \in \mathcal{E} - \{t\}\}$ and $T_{h}^1=\{(h, r^1,t^1) / (t^1, r^1, h) | r^1 \in \mathcal{R}$

\noindent $ - \{r\} , t^1 \in \mathcal{E} \}$; when $l>1$, $T_{hr/h}^i=\{(h^{i-1}, r^i, t^i$

\noindent $)/(t^i, r^i, h^{i-1})| h^{i-1} \in New(T_{hr/h}^{i-1}), r^{i} \in \mathcal{R}, t^i\in \mathcal{E} \}$, where "/" denotes "or", and $New(T_{hr/h}^{i-1})$ is the latest sampled entity set in $T_{hr/h}^{i-1}$.
For $T_{r}$, we solely consider distant triplets with $r$, i.e., $T_{r}=\{(h^\prime, r, t^\prime)|h^\prime, t^\prime \in \mathcal{E}-\{h, t\}\}$.

In the sampling process (e.g., $T_{hr/h}^i$ and $T_{r}$), we leverage Random Walk~\cite{ko2024subgraph} to select triplets based on the degree distribution of candidate entities, considering both out-degree and in-degree. Additionally, to control the size of the subgraph, we set the total number of sampled triplets to $m=m_{hr}+m_{h}+m_{r}$, where $m_{hr/h}$ and $m_{r}$ represent the sampling numbers of $T_{hr/h}$ and $T_{r}$, respectively. Note that if $|T_{hr/h}|<m_{hr/h}$, we select more distant triplets to ensure $m$.




\subsection{Improved Graph Transformer}
\label{iGTE_sec:}

\begin{figure*}[t]
  \centering
  \includegraphics[scale=0.35]{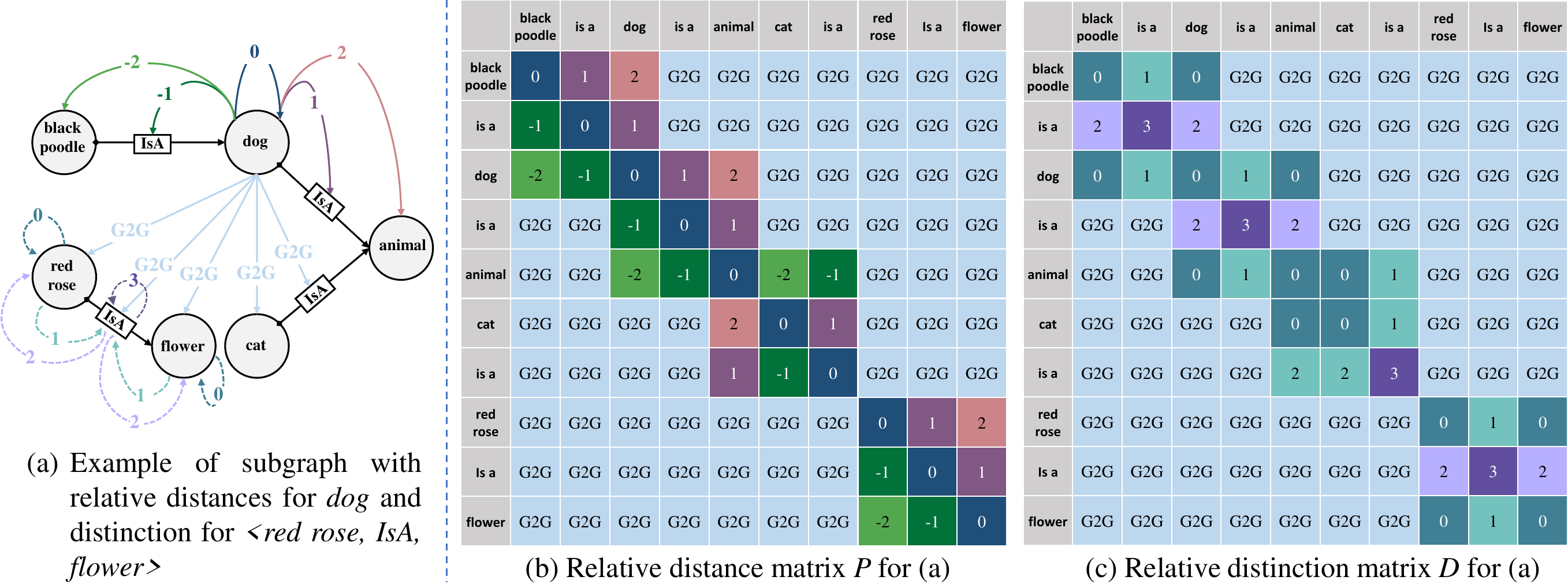}
  \caption{Example of subgraph preprocessing in iGT. We follow the construction strategy of the relative position matrix $P$ in $g$GLM~\cite{plenz-frank-2024-graph}. The relative distinction matrix $D$ differentiates entities and relations in iGT. Notably, it can be extended to $g$GLM, providing clear textual boundaries for entities and relations (see Appendix~\ref{app:P_D_gglm}). Also, entries with G2G are initialized to $+\infty$.}
  \label{fig1:P_D}
  \vspace*{-3ex}
\end{figure*}

In order to effectively encode $\mathcal{G}_{sub}(h,r,?)$, we propose an improved Graph Transformer (iGT). Concretely, we first introduce the three-word language and pre-compress the textual information of entities and relations. Given an entity $e$ and a relation $r$ from $\mathcal{G}_{sub}(h,r,?)$, their token embedding sequences of textual information take the following forms:
\begin{align}
    {\rm E}_{e} = [\bm{t}_e^1,\cdots, \bm{t}_e^{n_e}],
    {\rm E}_{r} = [\bm{t}_r^1,\cdots, \bm{t}_r^{n_r}], 
\end{align}
where $n_e$ and $n_r$ represent the lengths of the token sequences for textual information. Then, following~\cite{guo2024mkgl}, we draw on the pooling operator ${\rm Pool_{op}}()$ from PNA~\cite{corso2020principal} to compress ${\rm E}_{e}$ and ${\rm E}_{r}$, i.e.,
\begin{align}
    \bm{t}_{e} = {\rm Pool_{op}}({\rm E}_{e}),
    \bm{t}_{r} = {\rm Pool_{op}}({\rm E}_{r}), 
\end{align}
where $\bm{t}_{e}$ and $\bm{t}_{r}$ denote the textual token embedding of $e$ and $r$, respectively. By utilizing the pooling operator, we furnish embeddings for every entity and relation within $\mathcal{G}_{sub}(h,r,?)$.


Next, we construct a relative distance matrix $P$ with a global perspective for $\mathcal{G}_{sub}(h,r,?)$, following GLM, as shown in Fig.~\ref{fig1:P_D} (a) and (b). 
We regard triplets as three-word sentences, where each token represents an entity or a relation, and calculate their relative distances.
Moreover, the graph-to-graph (G2G) relative position (initialized as the parameter of the relative position for $+\infty$) can connect any token to other tokens, thereby enabling access to and learning of distant entities or relations.


Although $P$ achieves graph manipulation in a non-intrusive way, it fails to distinguish between entities and relations in $\mathcal{G}_{sub}(h,r,?)$, which may introduce confounding bias.
This is because in KG, entities represent real-world objects or concepts, while relations describe the interactions between entities~\cite{pan2024unifying}. 
To rectify this, we introduce a new relative distinction matrix $D$, which has the same shape as $P$ and shares G2G, as shown in Fig.~\ref{fig1:P_D}(c). Unlike $P$, $D$ aims to distinguish between entities and relations in the subgraph. 
To be specific, the relative positions between entities (i.e., entity-entity) are set to $0$ and populated into the corresponding ones in $D$.
Similarly, the positions for entity-relation, relation-entity, and relation-relation pairs are assigned the values of $1$, $2$, and $3$, respectively. 
Furthermore, we rewrite the Eq.~(\ref{eq:at_me}) of the attention mechanism as:
\begin{align}
    {\rm softmax}\left(\frac{QK^\top}{\sqrt{d}} + B_{PD}\right)V,
\end{align}
where $B_{PD}=\frac{1}{2}\left(f_1(P)+f_2(D)\right)$. Here, $f_1$ and $f_2$ are two different element-wise functions.
Compared with GLM, iGT focuses on the structural information of $\mathcal{G}_{sub}(h,r,?)$, bringing several benefits: it simplifies the complexity of positional encoding; handles larger subgraphs; and differentiates between entities and relations while treating them equitably. Importantly, iGT inherits GLM's non-invasive properties, circumventing the need to train the model from scratch, although the pooling operator may lose some textual information. 


With iGT, we can encode the subgraph $\mathcal{G}_{sub}(h,r,?)$, and the overall process is as follows:
\begin{align}
    & [h, r, ?,  \cdots] = {\rm ExTok}\left(\mathcal{G}_{sub}(h,r,?)\right), \notag \\
    &[\bm{t}_h, \bm{t}_r, \bm{t}_?,  \cdots] = {\rm Pool_{op}}\left({\rm Emb}([h, r, ?, \cdots])\right), \notag \\
    &[\tilde{\bm{t}}_h, \tilde{\bm{t}}_r, \tilde{\bm{t}}_?,  \cdots] = {\rm iGT}([\bm{t}_h, \bm{t}_r, \bm{t}_?,  \cdots], P, D), \notag
\end{align}
where ${\rm ExTok}$ is the Expanded Tokenizer, which integrates entities and relations as new tokens into the existing vocabulary. ${\rm Emb}$ denotes Embedding layer.
Of note, during training and prediction, we replace $?$ (to be predicted) with \textit{mask} token from the original Tokenizer.


\subsection{Subgraph-based Training Objective}
\label{sub_training_obj:sec}
In this paper, we frame the $(h, r, ?)$ prediction task as a multi-classification problem. To elaborate, we implement an MLP-based classification layer that takes $[\tilde{\bm{t}}_h, \tilde{\bm{t}}_r, \tilde{\bm{t}}_?]$ from iGT's final hidden layer as input, with its output dimension corresponding to the KG's total entity count $N$. 
Then, we compute classification probabilities through softmax activation and optimize using cross-entropy loss. Of note, prior to classification, we perform the pooling operation on $[\tilde{\bm{t}}_h, \tilde{\bm{t}}_r, \tilde{\bm{t}}_?]$. The process can be formulated as:
\begin{align}
    & \tilde{\bm{t}}_{(h,r,?)} = {\rm Pool_{op}}([\tilde{\bm{t}}_h, \tilde{\bm{t}}_r, \tilde{\bm{t}}_?]),  \label{pool_op_eq:} \\
    & \hat{\bm{t}}_{(h,r,?)} = {\rm softmax}({\rm MLP}(\tilde{\bm{t}}_{(h,r,?)})),  \\
    & \mathcal{L}_{ce} = -\log(\hat{\bm{t}}_{(h,r,?),l_{t}}), 
\end{align}
where $\hat{\bm{t}}_{(h,r,?),l_t}$ denotes the likelihood of entity $t$ being selected.

We don't use $\tilde{\bm{t}}_?$ alone as the classification input; instead, we opt for $[\tilde{\bm{t}}_h, \tilde{\bm{t}}_r, \tilde{\bm{t}}_?]$. This is because iGT encodes $\mathcal{G}_{sub}(h,r,?)$, in which $h$ may be shared by multiple triplets, and $r$ can also appear in several triplets (see Fig.~\ref{fig1:framework}).  Thus, the optimization objective based solely on $\tilde{\bm{t}}_?$ may not effectively address the prediction task $(h, r, ?)$. 
Also, according to Section~\ref{sub_extr:sec}, triplets in $T_{hr}^1$ feature the same head entity and relation as $(h, r, ?)$, such as $(h,r_1, {\color{red} h})$ and $(h,r_2, {\color{blue} h})$ (see Fig.~\ref{fig1:framework}). Hence, during prediction, ${\color{red} h}$ and ${\color{blue} h}$ can emerge as potential optimization targets requiring positive attention, whereas other entities, including $h$, warrant negative attention. 
To this end, we partition all entities in $\mathcal{G}_{sub}(h,r,?)$ (excluding $?$) into two sets: ${\rm Pos}$ and ${\rm Neg}$. ${\rm Pos}$ includes tail entities from all triplets in $T_{hr}^1$, while ${\rm Neg}$ comprises the remaining entities. The optimization objectives for ${\rm Pos}$ and ${\rm Neg}$ take the following forms:
\begin{align}
    & \mathcal{L}_{pos} = -\frac{1}{|{\rm Pos}|}\sum_{t^\prime\in {\rm Pos}}\log(\hat{\bm{t}}_{(h,r,t^\prime),l_{t^\prime}}),\\
    & \mathcal{L}_{neg} = -\frac{1}{|{\rm Neg}|}\sum_{t^\prime\in {\rm Neg}}\log(\hat{\bm{t}}_{(h,r,t^\prime),l_{t^\prime}}),
\end{align}
where $|{\rm Pos}|$ and $|{\rm Neg}|$ denote the number of entities in ${\rm Pos}$ and ${\rm Neg}$. 

Combining $\mathcal{L}_{ce}$, $\mathcal{L}_{pos}$, and $\mathcal{L}_{neg}$, the subgraph-based overall objective can be formalized as follows:
\begin{align}
\label{eq:overall_obj}
    \mathcal{L}  = \mathcal{L}_{ce} + \beta_{1} (\mathcal{L}_{pos}-\beta_{2} \mathcal{L}_{neg}),
\end{align}
where $\beta_{1}>0$ and $\beta_{2}>0$ are tunable hyperparameters. During training, to prevent $\mathcal{L}_{neg}$ from dominating excessively, we employ the following strategy to adjust $\beta_{2}$ adaptively:
\begin{align}
\label{beta2_cal:eq}
    \beta_{2}=\left\{\begin{array}{l}
    1,  \mathcal{L}_{pos} >  \mathcal{L}_{neg}, \\
    0.5 * \frac{\mathcal{L}_{pos}}{\mathcal{L}_{neg}},  \mathcal{L}_{pos} \leq \mathcal{L}_{neg}.
    \end{array}  \right. 
\end{align}

\begin{table*}[htbp]
  \centering
  \resizebox{2.0\columnwidth}{!}{
    \begin{tabular}{ccccc|cccc|cccc}
    \toprule
    \multirow{2}[4]{*}{Methods} & \multicolumn{4}{c|}{FB15k-237} & \multicolumn{4}{c|}{WN18RR}   & \multicolumn{4}{c}{Wikidata5M} \\
\cmidrule{2-13}          & MRR   & Hits@1 & Hits@3 & Hits@10 & MRR   & Hits@1 & Hits@3 & Hits@10 & MRR   & Hits@1 & Hits@3 & Hits@10 \\
    \midrule
    TransE & 0.279 & 0.198 & 0.376 & 0.441 & 0.243 & 0.043 & 0.441 & 0.532 & 0.392 & 0.323 & 0.432 & 0.509 \\
    RotatE & 0.338 & 0.241 & 0.375 & 0.533 & 0.476 & 0.428 & 0.492 & 0.571 & 0.403 & 0.334 & 0.441 & 0.523 \\
    HAKE  & 0.346 & 0.250 & 0.381 & 0.542 & 0.497 & 0.452 & 0.516 & 0.582 & 0.394 & 0.322 & 0.435 & 0.521 \\
    CompoundE & 0.350 & 0.262 & 0.390 & 0.547 & 0.492 & 0.452 & 0.510 & 0.570 & -     & -     & -     & - \\
    \midrule
    KG-BERT & -     & -     & -     & 0.420 & 0.216 & 0.041 & 0.302 & 0.524 & -     & -     & -     & - \\
    KG-S2S & 0.336 & 0.257 & 0.373 & 0.498 & 0.574 & 0.531 & 0.595 & 0.661 & -     & -     & -     & - \\
    CSProm-KG & 0.358 & 0.269 & 0.393 & 0.538 & 0.575 & 0.522 & \underline{0.596} & \underline{0.678} & 0.380  & 0.343 & 0.399 & 0.446 \\
    PEMLM-F & 0.355 & 0.264 & 0.389 & 0.538 & 0.556 & 0.509 & 0.573 & 0.648 & -     & -     & -     & - \\
    \midrule
    CompGCN & 0.355 & 0.264 & 0.390 & 0.535 & 0.479 & 0.443 & 0.494 & 0.546 & -     & -     & -     & - \\
    REP-OTE & 0.354 & 0.262 & 0.388 & 0.540 & 0.488 & 0.439 & 0.505 & 0.588 & -     & -     & -     & - \\
    KRACL & 0.360 & 0.266 & 0.395 & 0.548 & 0.527 & 0.482 & 0.547 & 0.613 & -     & -     & -     & - \\
    $g$GLM  & 0.321 & 0.241 & 0.342 & 0.486 & 0.290  & 0.304 & 0.395 & 0.487 & -     & -     & -     & - \\
    iGT (\textbf{ours}) & 0.364 & 0.283 & 0.411 & 0.566 & 0.534 & 0.496 & 0.536 & 0.617 & 0.397 & 0.342 & 0.428 & 0.526 \\
    \midrule
    GPT-3.5 & -     & 0.267 & -     & -     & -     & 0.212 & -     & -     & -     & -     & -     & - \\
    Llama-2-13B & -     & -     & -     & -     & -     & 0.315 & -     & -     & -     & -     & -     & - \\
    KICGPT & 0.412 & 0.327 & 0.448 & 0.554 & 0.549 & 0.474 & 0.585 & 0.641 & -     & -     & -     & - \\
    MPIKGC-S & 0.359 & 0.267 & 0.395 & 0.543 & 0.549 & 0.497 & 0.568 & 0.652 & -     & -     & -     & - \\
    KG-FIT  & 0.362 & 0.275 & -     & 0.572 & -     & -     & -     & -     & -     & -     & -     & - \\
    MKGL  & 0.415 & 0.325 & 0.454 & 0.591 & 0.552 & 0.500   & 0.577 & 0.656 & -     & -     & -     & - \\
    \midrule
    GLTW$_{1b}$ & 0.385 & 0.312 & 0.427 & 0.578 & 0.549 & 0.514 & 0.558 & 0.645 & 0.405 & 0.356 & 0.452 & 0.531 \\
    GLTW$_{3b}$ & \underline{0.427} & \underline{0.338} & \underline{0.462} & \underline{0.599} & \underline{0.578} & \underline{0.538} & 0.593 & 0.676 & \underline{0.429} & \underline{0.376} & \underline{0.476} & \underline{0.553} \\
    GLTW$_{7b}$ & \textbf{0.469} & \textbf{0.351} & \textbf{0.481} & \textbf{0.614} & \textbf{0.593} & \textbf{0.556} & \textbf{0.649} & \textbf{0.690}  & \textbf{0.457} & \textbf{0.414} & \textbf{0.506} & \textbf{0.587} \\
    \bottomrule
    \end{tabular}
    }%
    \caption{Performance comparison of various methods across different datasets. 
  Note that \textbf{bold} indicates the overall best performance, while \underline{underline} marks the second-best one.} 
  \label{tab1:compare_res}%
\end{table*}%

\subsection{Joint iGT and LLM}
\label{Join_igt_llm:sec}

We now combine iGT and LLM by fusing entity and relation embeddings. 
To be specific, we integrate the pooled embeddings of entity $h$ ($\bm{t}_h^{llm}$) and relation $r$ ($\bm{t}_r^{llm}$) from the LLM-based KG language prompt for $(h,r,?)$ with iGT's output embeddings [$\tilde{\bm{t}}_h, \tilde{\bm{t}}_{r}, \tilde{\bm{t}}_{r_1}, \cdots$], excluding $\tilde{\bm{t}}_{?}$.
The process (i.e., the Embedding Fusion Module) is defined as:
\begin{align}
    & \overline{\bm{t}}_r = {\rm Pool_{op}}([\tilde{\bm{t}}_{r}, \tilde{\bm{t}}_{r_1}, \cdots]), \label{t_r_r1:} \\
    & \bm{t}_h^{llm} \leftarrow (1-\lambda)\cdot \bm{t}_h^{llm} + \lambda \cdot {\rm Adapter}(\tilde{\bm{t}}_h), \label{t_llm_h:} \\
    & \bm{t}_r^{llm} \leftarrow (1-\lambda)\cdot \bm{t}_r^{llm} + \lambda \cdot {\rm Adapter}(\overline{\bm{t}}_r), \label{t_llm_r:}
\end{align} 
where $\lambda \in [0,1]$, and ${\rm Adapter}$ aims to align embedding dimensions. 
The selection of ${\rm Adapter}$ is flexible. 
In practice, following~\cite{zhu2023minigpt}, we implement ${\rm Adapter}$ as a simple projection layer. 
Notably, we pass the pooled relation embeddings $[\tilde{\bm{t}}_{r}, \tilde{\bm{t}}_{r_1}, \cdots]$ from $\mathcal{G}_{sub}(h,r,?)$ to the LLM, enabling it to capture global KG structural information.

Then, we incorporate the embedding vector $\bm{t}_{hr}^{llm}$ from the last token of the LLM's final hidden layer into the classification layer as follows:
\begin{align}
\label{concat_eq:}
    \tilde{\bm{t}}_{(h,r,?)} \leftarrow {\rm Concat}(\bm{t}_{hr}^{llm}, \tilde{\bm{t}}_{(h,r,?)}),
\end{align}
where $\tilde{\bm{t}}_{(h,r,?)}$ is derived from Eq.~(\ref{pool_op_eq:}). Similarly, all positive and negative triplets constructed in Section~\ref{sub_training_obj:sec} are combined with $\bm{t}_{hr}^{llm}$ in the same manner (see Fig.~\ref{fig1:framework}). Of note, the input dimension of the MLP classification layer changes accordingly through Eq.~(\ref{concat_eq:}).

\section{Experiments}

\subsection{Experimental Settings}

\textbf{Datasets.}
We evaluate different methods on three widely used KG datasets, including FB15k-237~\cite{toutanova2015representing}, WN18RR~\cite{dettmers2018convolutional}, and Wikidata5M~\cite{vrandevcic2014wikidata}, for the link prediction task.
We detail these datasets in Table~\ref{tab:data_statistical} from Appendix~\ref{App:com_exper_settings}.

\textbf{Baselines.}
To assess the effectiveness of our methods, we follow~\cite{plenz-frank-2024-graph} by adopting the bidirectional encoder of T5-base as the base Pre-trained Language Model (PLM) for \textbf{iGT}. 
Meanwhile, we choose three LLMs with varying sizes for GLTW: Llama-3.2-1B/3B-Instruct~\cite{dubey2024llama} and Llama-2-7b-chat~\cite{touvron2023llama}. For clarity, we denote GLTW with different LLMs as \textbf{GLTW$_{1b/3b/7b}$}. Also, we compare GLTW and iGT against numerous embedding-based, text-based, GNN/GT-based and LLM-based baselines.
The embedding-based baselines include TransE~\cite{bordes2013translating}, RotatE~\cite{sun2019rotate}, HAKE~\cite{zhang2020learning}, and CompoundE~\cite{ge2023compounding}.
The text-based baselines encompass KG-BERT~\cite{yao2019kg}, KG-S2S~\cite{chen2022knowledge}, CSProm-KG~\cite{chen2023dipping}, and PEMLM-F~\cite{qiu2024joint}.
The GNN/GT-based baselines cover CompGCN~\cite{vashishth2019composition}, REP-OTE~\cite{wang2022simple}, and KRACL~\cite{tan2023kracl} (based on GNN), as well as $g$GLM~\cite{plenz-frank-2024-graph} (based on GT). Note that $g$GLM and iGT are trained on identical subgraphs.
The LLM-based baselines comprise GPT-3.5-Turbo with one-shot ICL (marked as GPT-3.5)~\cite{zhu2024llms}, Llama-2-13B+Struct (marked as Llama-2-13B)~\cite{yao2023exploring}, KICGPT~\cite{wei2024kicgpt}, MPIKGC-S~\cite{xu2024multi}, KG-FIT~\cite{jiang2024kg}, and MKGL~\cite{guo2024mkgl}.

\textbf{Configurations.}
In all experiments, unless otherwise specified, we default to setting $l=2$ and $\overline{m}=m_{hr}=m_{h}=m_{r}=m/3=5$ for subgraph sampling. Meanwhile, we set $\lambda=0.5$ and $\beta_1=0.5$. 
Of note, $\beta_2$ is adaptively calculated based on Eq.~(\ref{beta2_cal:eq}). Also, we assess performance by leveraging the Mean Reciprocal Rank (MRR) of target entities and the percentage of target entities ranked in the top $k$ ($k=1,3,10$), referred to as Hits@$k$.
Due to space limitations, the complete experimental settings are provided in Appendix~\ref{App:com_exper_settings}.

\subsection{Results Comparison}
We compare the proposed methods with various KGC baselines on FB15k-237, WN18RR, and Wikidata5M, with the results shown in Table~\ref{tab1:compare_res}. The results indicate that: 
1) GLTW$_{7b}$ consistently outperforms all competitors across all metrics, achieving overall gains of $8.5\%$ in MRR, $6.9\%$ in Hits@1, $10.2\%$ in Hits@3, and $6.1\%$ in Hits@10 compared to the second-best results (mostly from GLTW$_{3b}$). Meanwhile, GLTW's performance improves as the LLM size increases. These results demonstrate that GLTW effectively captures the characteristics of entities and relations in KGs and leverages the rich knowledge in LLMs to enhance prediction accuracy.
2) GLTW$_{3b}$ beats Llama-2-7b-based baseline MKGL (the most comparable method) on all metrics for FB15k-237 and WN18RR, with further improvements achieved by GLTW$_{7b}$. 
We attribute GLTW's advantage to its effective encoding of both local and global structural information of KGs, tailoring a suitable objective function for training subgraphs, and enabling LLMs to perceive structural information and effectively participate in entity prediction.
3) The proposed iGT consistently outstrips other GT/GNN-based baselines on FB15k-237 and WN18RR, while $g$GLM uniformly lags behind others. A detailed analysis is provided in the Ablation Study.

\subsection{Ablation Study}
In this section, we carefully demonstrate the efficacy and
indispensability of the core modules and key parameters in our
methods on FB15k-237 and WN18RR.
\begin{table}[htbp]
  \centering
  \resizebox{1.0\columnwidth}{!}{
    \begin{tabular}{lcccc|cccc}
    \toprule
    \multirow{2}[4]{*}{Method} & \multicolumn{4}{c|}{FB15k-237} & \multicolumn{4}{c}{WN18RR} \\
\cmidrule{2-9}          & MRR   & Hits@1 & Hits@3 & Hits@10 & MRR   & Hits@1 & Hits@3 & Hits@10 \\
    \midrule
    \rowcolor{Gray}GLTW$_{1b}$ & \textbf{0.385} & \textbf{0.312} & \textbf{0.427} & \textbf{0.578} & \textbf{0.549} & \textbf{0.514} & \textbf{0.558} & \textbf{0.645} \\
    -w/o. iGT & 0.108 & 0.082 & 0.177 & 0.303 & 0.205 & 0.157 & 0.261 & 0.413 \\
    -w/o. FT for LLM & 0.379 & 0.291 & 0.397 & 0.572 & 0.539 & 0.509 & 0.545 & 0.629 \\
    \midrule
    \rowcolor{Gray}GLTW$_{3b}$ & \textbf{0.427} & \textbf{0.338} & \textbf{0.462} & \textbf{0.599} & \textbf{0.578} & \textbf{0.538} & \textbf{0.593} & \textbf{0.676} \\
    -w/o. iGT & 0.171 & 0.145 & 0.191 & 0.325 & 0.287 & 0.222 & 0.309 & 0.439 \\
    -w/o. FT for LLM & 0.411 & 0.323 & 0.445 & 0.587 & 0.552 & 0.529 & 0.567 & 0.663 \\
    \midrule
    \rowcolor{Gray}GLTW$_{7b}$ & \textbf{0.469} & \textbf{0.351} & \textbf{0.481} & \textbf{0.614} & \textbf{0.593} & \textbf{0.556} & \textbf{0.649} & \textbf{0.690} \\
    -w/o. iGT & 0.207 & 0.184 & 0.236 & 0.366 & 0.394 & 0.309 & 0.357 & 0.462 \\
    -w/o. FT for LLM & 0.438 & 0.343 & 0.465 & 0.607 & 0.568 & 0.538 & 0.612 & 0.677 \\
    \midrule
    -w/o LLM (i.e., iGT) & 0.364 & 0.283 & 0.411 & 0.566 & 0.534 & 0.496 & 0.536 & 0.617 \\
    \bottomrule
    \end{tabular}
    }%
    \caption{Impact of each component for GLTW.}
  \label{tab:ablation_non_FT}%
\end{table}%

\textbf{Necessity of each component for GLTW.} 
To investigate the impact of iGT and LLMs on the performance for GLTW, we establish three control baselines: training iGT alone (w/o. LLM), fine-tuning LLMs alone (w/o. iGT), and using GLTW without fine-tuning LLMs (w/o. FT for LLM). For LLM fine-tuning alone, we input the KG language prompt and use the embedding vector of the last token from the final hidden layer as input to the classification layer. 
We report the results in Table~\ref{tab:ablation_non_FT}.
One can observe that iGT and LLMs exhibit significant performance drops compared to GLTW with different-sized LLMs.
Specifically, iGT sees average declines of 5.1\%, 4.5\%, 5.5\%, and 4.2\% in MRR, Hits@1, Hits@3, and Hits@10, respectively, while LLMs experience average drops of 27.2\%, 25.2\%, 27.3\%, and 24.9\% in these metrics.
Notably, GLTW without fine-tuning LLMs still surpasses both iGT and 
LLMs. This confirms that combining iGT and LLMs enhances entity prediction, consistent with prior works~\cite{qiu2024joint, zhang2024making}. Meanwhile, our proposed joint strategy effectively unlocks the LLM's potential for the link prediction task. 
Additionally, iGT consistently trumps all LLMs, underscoring the critical importance of relevant KG information and a well-designed training objective for performance improvements.

\begin{table}[htbp]
  \centering
  \resizebox{1.0\columnwidth}{!}{
    \begin{tabular}{lcccc|cccc}
    \toprule
    \multirow{2}[4]{*}{Method} & \multicolumn{4}{c|}{FB15k-237} & \multicolumn{4}{c}{WN18RR} \\
\cmidrule{2-9}          & Hits@1 & A.IT($\downarrow$) & A.IL($\downarrow$) & A.BBR($\downarrow$) & Hits@1 & A.IT($\downarrow$) & A.IL($\downarrow$) & A.BBR($\downarrow$) \\
    \midrule
    \rowcolor{Gray} iGT   & \textbf{0.283} & \textbf{16}    & \textbf{38.43} & \textbf{0.00}     & \textbf{0.496} & \textbf{16}    & \textbf{44.16} & \textbf{0.00} \\
    -w/o. $\mathcal{L}_{pos}$ & 0.263 & -     & -     & -     & 0.471 & -     & -     & - \\
    -w/o. $\mathcal{L}_{neg}$ & 0.274 & -     & -     & -     & 0.484 & -     & -     & - \\
    -w/o. $\mathcal{L}_{pos}$ \& $\mathcal{L}_{neg}$  & 0.243 & -     & -     & -     & 0.430  & -     & -     & - \\
    \hdashline[2pt/3pt]
    -w/o. $D$ & 0.254 & -     & -     & -     & 0.453 & -     & -     & - \\
    \midrule
    \rowcolor{Gray} $g$GLM  & 0.241 & 14.9  & 313.57 & 0.01  & 0.304 & 9.45  & 448.21 & 0.34 \\
    -w. $D$ & 0.267 & -     & -     & -     & 0.346 & -     & -     & - \\
    \bottomrule
    \end{tabular}%
    }
    \caption{Utility of $D$ and various parts in Eq.~(\ref{eq:overall_obj}) for iGT, as well as iGT vs. $g$GLM}
  \label{tab:igt_gglm}%
\end{table}%

\textbf{Utility of $\mathcal{L}$ and $D$.} 
We delve into the subgraph-based training objective $\mathcal{L}$ and the relative discrimination matrix $D$ by leveraging iGT. Thereafter, due to space limitations, we only report the values of Hits@1.
For the \textbf{former}, we perform the leave-one-out test to explore the individual contributions of $\mathcal{L}_{pos}$ and $\mathcal{L}_{neg}$ to iGT, and further display the test results by simultaneously discarding them.
As shown in Table~\ref{tab:igt_gglm}, removing
either $\mathcal{L}_{pos}$ or $\mathcal{L}_{neg}$ adversely affects the performance of iGT.
In addition, the absence of both losses further worsens the decline of Hits@1, demonstrating that $\mathcal{L}_{pos}$ and $\mathcal{L}_{neg}$ are vital for training subgraphs.
Interestingly, we observe that removing $\mathcal{L}_{pos}$ has a more pronounced negative impact than removing $\mathcal{L}_{neg}$.
The empirical results indicate that in subgraph-based training, the construction of positive and negative triplets (i.e., PT and NT) is crucial for capturing structural information in KGs. 
For the \textbf{latter}, Table~\ref{tab:igt_gglm} reveals that removing $D$ from iGT decreases the Hits@1 value by $2.9\%$ and $4.3\%$ on FB15k-237 and WN18RR, respectively. Importantly, extending $D$ to $g$GLM improves Hits@1 value by $2.6\%$ and $4.2\%$ on these datasets. This suggests that $B_{DP}$ enhances the relative positional encoding of entities and relations for subgraphs compared to $B_P$.

Notably, we illustrate the encoding strategy of $D$ in $g$GLM, as shown in Fig.~\ref{fig:p_D_gGLM}(c) of Appendix~\ref{app:P_D_gglm}.
Essentially, $D$ introduces boundaries to the textual descriptions of entities and relations in subgraphs, thereby augmenting the PLM's perception of triples within KG.
Additionally, Table~\ref{tab:igt_gglm} records the three metrics during training for iGT and $g$GLM: average input triplets (\textbf{A.IT}), average input length (\textbf{A.IL}), and average tokens beyond the bucket range (\textbf{A.BBR}). The results show that iGT retains more input KG information than $g$GLM in terms of A.IT and A.BBR, especially on the WN18RR dataset. In contrast, the A.IL of $g$GLM is significantly higher than that of iGT, implying a higher computational cost for $g$GLM.
Therefore, we speculate that $g$GLM's underperformance in the link prediction task may be due to: 1) the lack of clear boundaries for entities and relations; 2) significant information loss when handling KGs with lengthy textual descriptions; and 3) potential bias introduced by focusing more on entities or relations with longer textual descriptions in each triplet.

\begin{figure}[h]
  \centering
  \begin{subfigure}{1.0\linewidth}
    \centering
    \includegraphics[width=\linewidth]{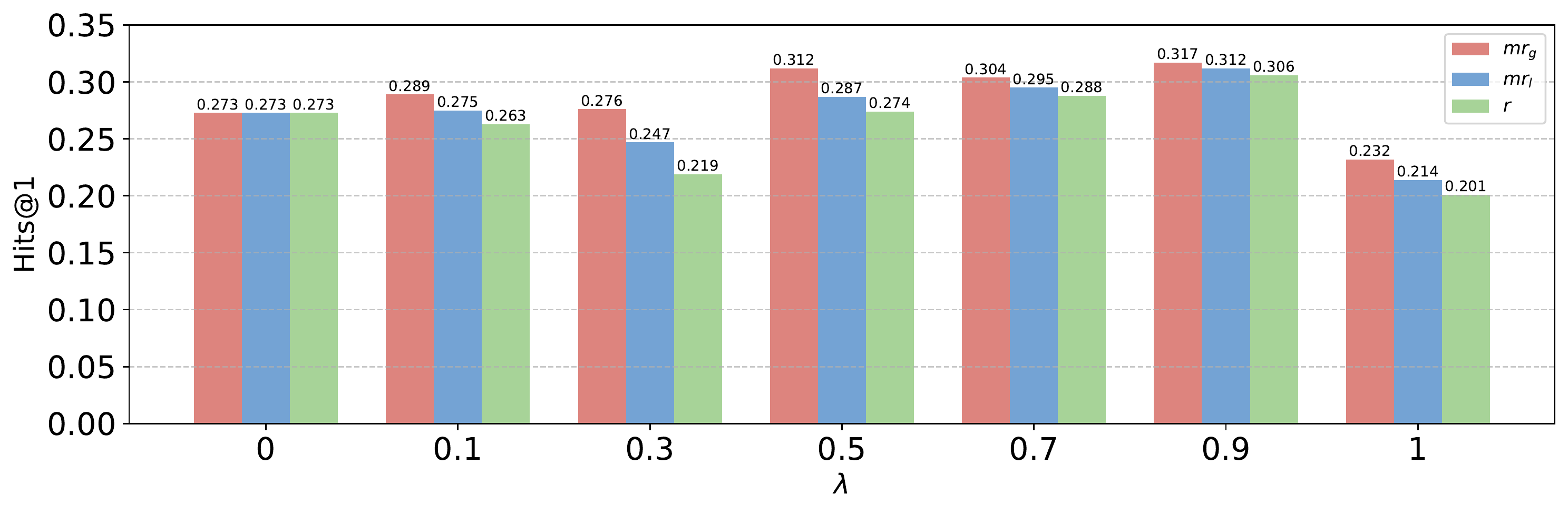}
        \caption{FB15k-237}
  \end{subfigure}
  \begin{subfigure}{1.0\linewidth}
    \centering
    \includegraphics[width=\linewidth]{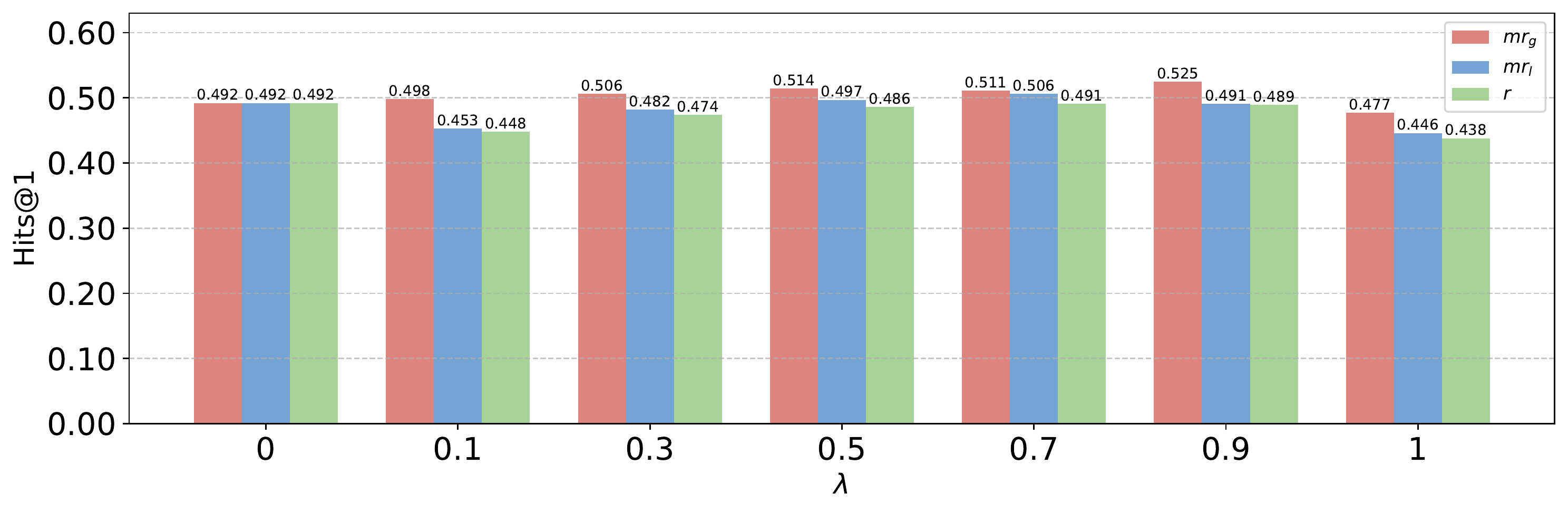}
        \caption{WN18RR}
  \end{subfigure}
  \caption{Hits@1 with varying $\lambda$ over FB15k-237 and WN18RR.} 
  \label{fig:lambda}
  \vspace*{-2ex}
\end{figure}

\textbf{Varying $\lambda$.}
We explore the impacts of $\lambda$ based on GLTW$_{1b}$ and select it from $\{0.0, 0.1, 0.3, 0.5, 0.7, 0.9, 1.0\}$. 
Additionally, we compare the performance of various relation embeddings: those appearing in triplets from $T_{hr}$ and $T_{h}$ (i.e., $\overline{\bm{t}}_r = {\rm Pool_{op}}([\tilde{\bm{t}}_{r}, \tilde{\bm{t}}_{r_1}, \cdots])$, marked as $mr_l$), those present in ones from $T_{hr}$, $T_{h}$ and $T_{r}$ (i.e., $\overline{\bm{t}}_r = {\rm Pool_{op}}([\tilde{\bm{t}}_{r}, \cdots, \tilde{\bm{t}}_{r_3}, \cdots])$, marked as $mr_g$), and the single relation in the target triplet (i.e., $\overline{\bm{t}}_r = \tilde{\bm{t}}_{r}$, marked as $r$), as shown in Eq.~(\ref{t_r_r1:}).
Fig.~\ref{fig:lambda} shows that GLTW with $\lambda \notin \{0.0, 1.0\}$ consistently dominates that with $\lambda \in \{0.0, 1.0\}$ in terms of Hits@1. Moreover, the performance with $\lambda=0$ is superior to that with $\lambda=1$. Notably, the Hits@1 values for $mr_{l/g}$ and $r$ are identical when $\lambda=0$, as the LLM only takes the KG language prompt as input, independent of them. These results indicate that our proposed combination of iGT and LLM effectively improves link prediction. 
Furthermore, we find that both $mr_g$ and $mr_l$ consistently outperform $r$ w.r.t. Hits@1, with $mr_g$ uniformly surpassing $mr_l$. 
This demonstrates that incorporating local structural information  (i.e., $T_{hr}$ and $T_{h}$) from KG into the training process improves the prediction accuracy for target entities, while adding global structural information (i.e., $T_r$) further boosts performance significantly.




\begin{figure}[h]
  \centering
  \begin{subfigure}{1.0\linewidth}
    \centering
    \includegraphics[width=\linewidth]{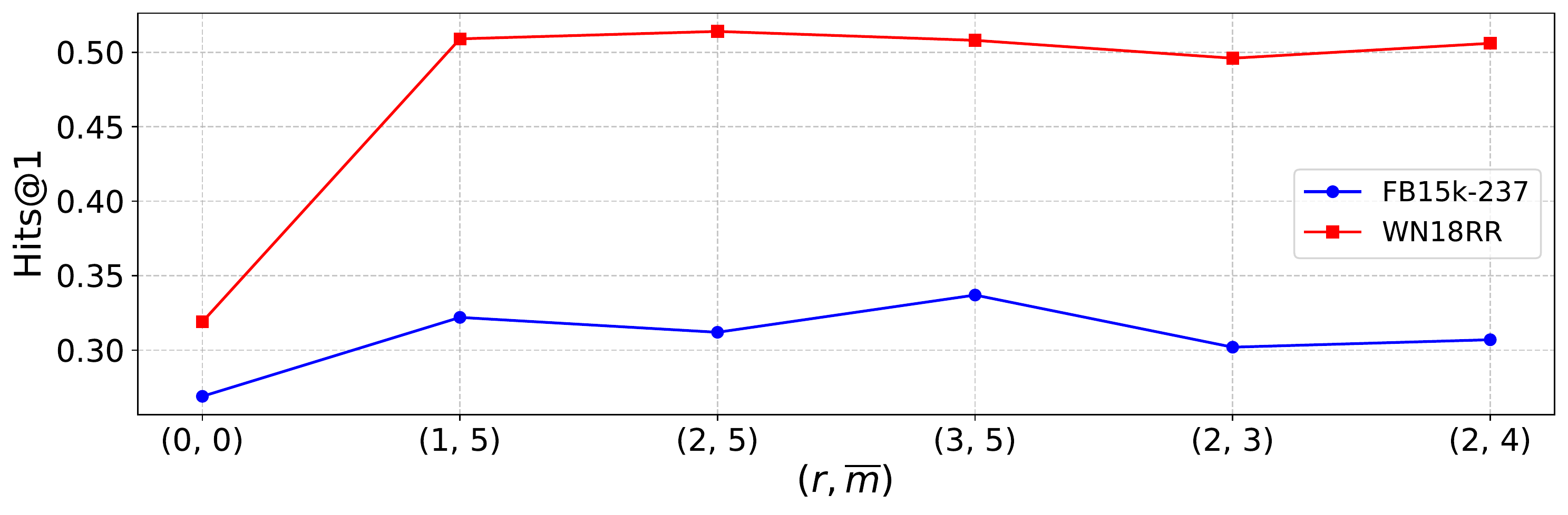}
        \caption{ }
  \end{subfigure}
  \begin{subfigure}{1.0\linewidth}
    \centering
    \includegraphics[width=\linewidth]{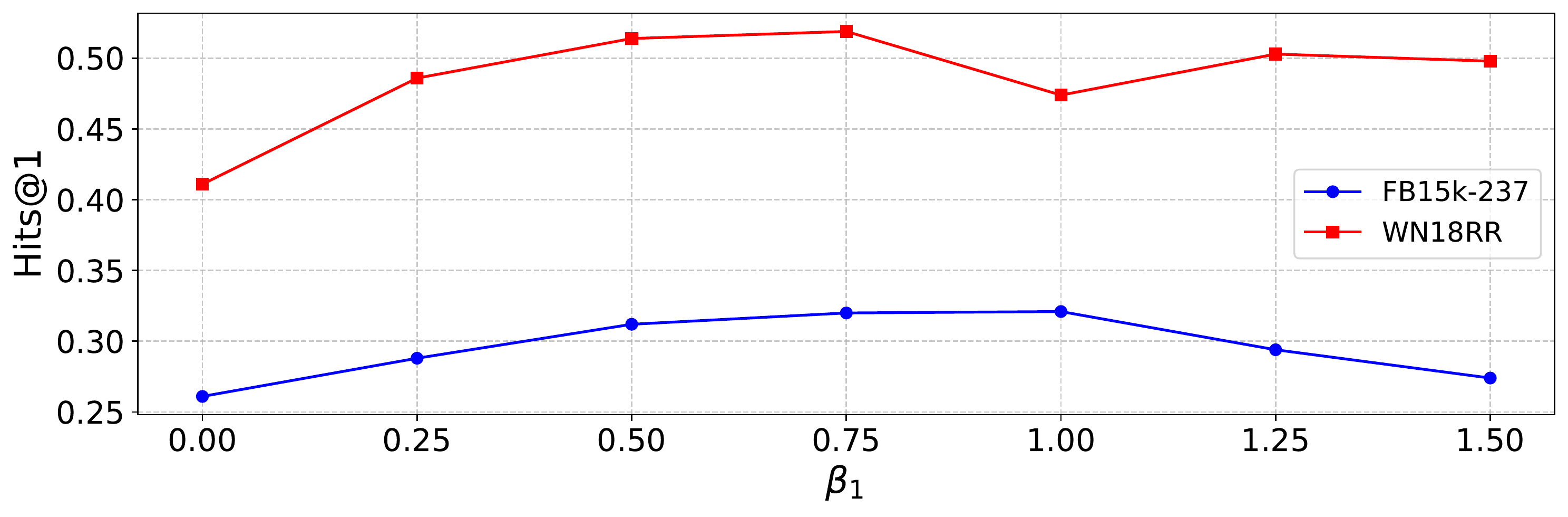}
        \caption{ }
  \end{subfigure}
  \caption{Hits@1 with varying $(r, \overline{m})$ and $\beta_1$ over FB15k-237 and WN18RR.} 
  \label{fig:r_m_beta}
  \vspace*{-2ex}
\end{figure}

\textbf{Varying $(r,\overline{m})$ and $\beta_1$.} 
We look into the effects of the parameters$(r, \overline{m})$, which control subgraph shape, and the constraint parameter $\beta_1$ for $\mathcal{L}$ using GLTW$_{1b}$. First, we set $(r, \overline{m})$ to values in $\{(0,0), (1,5), (2,5), (3,5), (2,3), (2,4)\}$ and report the results in Fig.~\ref{fig:r_m_beta}(a). Here, $(0,0)$ means that the subgraph contains only the target triplet $(h, r, ?)$. One can see that GLTW$_{1b}$ with $(r, m)=(0,0)$ underperforms other cases w.r.t. Hits@1, indicating that incorporating graph structure information significantly enhances entity prediction.
Furthermore, when $r=2$, GLTW$_{1b}$'s Hits@1 value improves as $\overline{m}$ increases, suggesting that moderately enlarging the subgraph scale intensifies performance. However, when $\overline{m}=5$, GLTW$_{1b}$'s performance does not monotonically improve with increasing $r$, highlighting the significantly impact of the subgraph sampling strategy on GLTW$_{1b}$'s performance for a given $\overline{m}$.
For $\beta_1$, we select values from $\{0.0, 0.25, 0.5, 0.75, 1.0, 1.25, 1.5\}$, as shown in Fig.~\ref{fig:r_m_beta}(b). We observe that the Hits@1 score of GLTW$_{1b}$  initially rises and then declines as $\beta_1$ increases. This indicates that the optimal $\beta_1$ depends on the scenario and requires case-specific tuning.

\section{Conclusion}
In this paper, we propose a novel method, GLTW, which aims to encode the structural information of KGs and integrate it with LLMs to
enhance KGC performance. Specifically, we formulate an improved graph transformer (iGT) that effectively encodes subgraphs with both local and global structural information and inherits the characteristics of language models, thus circumventing the need for training from scratch. Also, we develop a subgraph-based multi-classification training objective that treats all entities within KG as classification objects to improve learning efficiency. Importantly, we combine iGT with an LLM that takes KG language prompt as input.
Finally, we conduct extensive experiments to verify the superiority of GLTW.

\section*{Limitations}
Although empirical experiments have confirmed the effectiveness of the proposed GLTW, it still has two main limitations. \textbf{First}, training GLTW involves distinct original vocabularies for the T5 and Llama series, resulting in separate vocabularies for iGT and LLM. We speculate that well-trained GLTW on a unified vocabulary could further enhance its performance, but this would require training the models from scratch. \textbf{Second}, our proposed method uses pooling operations from PNA to compress textual information, which inevitably leads to some information loss. However, a key advantage of pooling operations is that they do not introduce new parameters requiring optimization. Even when training resources are limited and LoRA technology~\cite{hu2021lora} is drawn to reduce memory 
consumption, the additional trainable parameters are negligible. Therefore, it is crucial to develop pooling operations that minimize such information loss, which we leave to future work.

\section*{Ethical Considerations}
In this paper, all research and experiments utilize publicly available open-source datasets and models. 
We will release our code to support open research. 
Therefore, there is no ethical consideration in this paper.

\section*{Acknowledgments}
This work is supported by the National Science and Technology Major Project (2020AAA0106502), the National Natural Science Foundation of China (No.T2341003) and a grant from the Guoqiang Institute, Tsinghua University.

\bibliography{custom}

\clearpage
\appendix

\section*{Appendix}


\section{Related Work}
\label{Related_work:}

\textbf{Knowledge Graph Completion (KGC)} has evolved over the past decade and is a key task in the field of KGs. 
Mainstream KGC methods roughly fall into two groups: embedding-based and text-based methods.
Embedding-based methods~\cite{bordes2013translating, lin2015learning, sun2019rotate, balavzevic2019tucker} generate low-dimensional vectors for entities and relations and optimize various loss functions with the goal of $h+r \sim t$ to predict missing triplets. 
Although simple and effective, these methods neglect the extensive textual information in KGs and struggle to handle entities and relations not encountered during training.
On the other hand, text-based methods~\cite{yao2019kg, zhang2020pretrain, wang2022simkgc, liu2022know, wang2022language, yang2024knowledge} utilize the textual descriptions of entities and relations as input to pre-trained language models (PLMs) and introduce contrastive learning to enhance discriminative ability. 
However, these methods lack the inherent structural knowledge of KGs.
Consequently, some efforts~\cite{wang2021structure, chen2023dipping, he2024mocosa, yang2024knowledge, qiu2024joint} combine embedding- and text-based KGC methods, achieving improved performance.

\textbf{Graph Transformers} (GTs) are essentially a special type of GNN~\cite{bronstein2021geometric} and are gaining increasing attention in multiple application fields~\cite{chen2024survey}. In KGC, some studies~\cite{schlichtkrull2018modeling, vashishth2019composition, nathani2019learning, chen2020hitter, wang2022simple, tan2023kracl, galkin2023towards} leverage GNNs to encode structural information in KGs to train embeddings for entities and relations, while initializing them with semantic embeddings via PLMs. Recently, some efforts have explored applying GTs to KG-related tasks, e.g., graph-to-text generation~\cite{schmitt2020modeling, li2024unifying} and relation classification~\cite{plenz-frank-2024-graph}. However, they either train their models from scratch or split entities and relations into multiple tokens to construct complex positional encoding matrices.
For example, GLM~\cite{plenz-frank-2024-graph} is a graph transformer that fuses textual and structural information, enabling sequence PLMs to perform graph inference while maintaining their original ability.

However, GLM restricts the relative distance of individual triplets to between $0$ and $32$, which limits the processing of entities or relations with longer textual information. 
For instance, only $12.5\%$ of triplets in WN18RR (using the T5 tokenizer) fall within this distance range. 
Intuitively, the constraints of integrating textual and structural information also limit the size of processable subgraphs.
In addition, the attention mechanism may exhibit bias towards entities or relations with longer texts in each triplet. 
In this paper, we borrow the positional encoding strategy from GLM but shift our focus towards subgraph structural information while preserving GLM's strengths. We introduce a novel relative distinction matrix to achieve differentiated yet equal treatment of entities and relations in triplets. 
Our work is also the first to apply GT to the link prediction task.

\textbf{KGC with LLMs.} 
LLMs have been explored by researchers for various tasks due to their powerful emergent capabilities~\cite{luo2024let, yu2024automated, tan2024llm, si2025aligning, liu2025exploring}. 
Recently, LLMs are deemed highly promising in the realm of KGC and have garnered extensive attention~\cite{ren2024survey, pan2024unifying}.
For instance, \cite{yao2023exploring, zhu2024llms, wei2024kicgpt, li2024contextualization, xu2024multi} directly perform KGC via ICL or enhance textual information in KGs to improve text-based methods. However, these methods overlook the inherent structural information of KGs, leaving LLMs unable to perceive structural knowledge. To tackle this, \cite{zhang2024making, liu2024can, yang2024enhancing} integrate structural information with LLMs to boost KGC performance. Recently, MKGL~\cite{guo2024mkgl} enables LLMs to proficiently grasp entities and relations of KGs through three-word language, but how to make LLMs perceive graph information and improve the link prediction task remains an open problem.
Going beyond the aforementioned methods, there are a handful of recent studies~\cite{li2024cosign, xue2024unlock, jiang2024kg} on leveraging LLMs for KGC.

\section{Complete Experimental Settings}
\label{App:com_exper_settings}

\textbf{Datasets.}
We evaluate different methods with three widely used KG datasets, namely FB15k-237~\cite{toutanova2015representing}, WN18RR~\cite{dettmers2018convolutional}, Wikidata5M~\cite{vrandevcic2014wikidata}, for link prediction.
We detail the said datasets in Table~\ref{tab:data_statistical}.
Specifically, FB15k-237 is a curated dataset extracted from the Freebase~\cite{bollacker2008freebase} knowledge graph, covering knowledge across various domains, including movies, sports events, awards, and tourist attractions.
WN18RR is a well-known dataset built from WordNet~\cite{miller1995wordnet}, designed for knowledge graph research. It extracts a selection of lexical items and semantic relationships, covering a rich array of English words and their connections, such as synonyms, antonyms, and hierarchical relationships.
Wikidata5M~\cite{vrandevcic2014wikidata} is a large-scale KG dataset that integrates Wikidata and Wikipedia pages. Each entity in the dataset corresponds to a Wikipedia page, enabling it to support link prediction task for unseen entities. It follows the Wikidata identifier system, with entities prefixed by “Q” and relations by “P.” Additionally, the dataset provides a text corpus aligned with the KG structure.

\textbf{Baselines.}
To assess the effectiveness of our methods, we follow~\cite{plenz-frank-2024-graph} by using the bidirectional encoder of T5-base as the base PLM for \textbf{iGT}. 
Here, $P$ and $D$ are bucketed and mapped to $B_{PD}$ respectively, with sharing across layers.
Meanwhile, we choose three LLMs with different sizes for GLTW: Llama-3.2-1B/3B-Instruct~\cite{dubey2024llama}, and Llama-2-7b-chat~\cite{touvron2023llama}. For differentiation, we denote GLTW with different LLMs as \textbf{GLTW$_{1b/3b/7b}$}.

Also, we compare proposed GLTW and iGT against numerous embedding-based, text-based, GNN/GT-based and LLM-based baselines, which are the most relevant methods to our work.
The embedding-based baselines include TransE~\cite{bordes2013translating}, RotatE~\cite{sun2019rotate}, HAKE~\cite{zhang2020learning}, and CompoundE~\cite{ge2023compounding}.
The text-bsed baselines encompass KG-BERT~\cite{yao2019kg}, KG-S2S~\cite{chen2022knowledge}, CSProm-KG~\cite{chen2023dipping}, and PEMLM-F~\cite{qiu2024joint}.
The GNN/GT-based baselines cover CompGCN~\cite{vashishth2019composition}, REP-OTE~\cite{wang2022simple}, and KRACL~\cite{tan2023kracl} (based on GNN), as well as $g$GLM~\cite{plenz-frank-2024-graph} (based on GT). Note that $g$GLM and iGT are trained on the same sampled subgraphs.
The LLM-based baselines feature GPT-3.5-Turbo with one-shot ICL (marked as GPT-3.5)~\cite{zhu2024llms}, KG-Llama-2-13B+Struct (marked as Llama-2-13B)~\cite{yao2023exploring}, KICGPT~\cite{wei2024kicgpt}, MPIKGC-S~\cite{xu2024multi}, KG-FIT~\cite{jiang2024kg}, and MKGL~\cite{guo2024mkgl}.

\begin{table}[htbp]
  \centering
  \resizebox{1.0\columnwidth}{!}{
    \begin{tabular}{l|c|c|c|c|c}
    \toprule
    Dataset & \multicolumn{1}{l|}{\#Ent} & \multicolumn{1}{l|}{\#Rel} & \multicolumn{1}{l|}{\#Train} & \multicolumn{1}{l|}{\#Valid} & \multicolumn{1}{l}{\#Test} \\
    \midrule
    WN18RR & 40943 & 11    & 86835 & 3034  & 3134 \\
    FB15k-237 & 14541 & 237   & 272115 & 17535 & 20466 \\
    Wikidata5M & 4594485 & 822   & 20614279 & 5133  & 5163 \\
    \bottomrule
    \end{tabular}
    }%
    \caption{Statistics of the Datasets. Columns 2-6 represent the number of entities, relations, triples in the training set, triples in the validation set, and triples in the test set, respectively.}
  \label{tab:data_statistical}
\end{table}%

\textbf{Configurations.}
In all experiments, unless otherwise specified, we default to setting $l=2$ and $\overline{m}=m_{hr}=m_{h}=m_{r}=m/3=5$ for subgraph sampling. Meanwhile, we set $\lambda=0.5$ and $\beta_1=0.5$ by default. Note that $\beta_2$ is adaptively calculated based on Eq.~(\ref{beta2_cal:eq}). Also, we assess performance by leveraging the Mean Reciprocal Rank (MRR) of target entities and the percentage of target entities ranked in the top $k$ ($k=1,3,10$), referred to as Hits@$k$.

During training, we assign distinct training schedules to different modules to fully capture the knowledge in the KG datasets. These modules include iGT Encoder, LLM, Adapter and Classification Layer. Notably, we may also train the pooling operators. When training resources are limited, we follow~\cite{guo2024mkgl} by drawing on LoRA technology~\cite{hu2021lora} to mitigate memory consumption. For ease of description, we divide the training modules of GLTW into three parts: iGT Encoder, LLM, and the remaining modules (referred to as "Other Modules").
Specifically, for FB15k-237 and WN18RR, we set the number of training epochs to $10$ and the gradient accumulation steps to $4$. For Wikidata5M, we set the number of training epochs to $2$ and the gradient accumulation steps to $10$. In all experiments, we used a linear learning rate schedule and the AdamW optimizer. For iGT Encoder, LLM, and Other Modules, we set the learning rates to $0.0001$, $0.00001$, and $0.001$, respectively, with warm-up rates (i.e., the proportion of warm-up steps to total training steps) of $0.02$, $0.04$, and $0.01$.
Given that we used three different-sized LLMs, during training, we set the batch size per device to 16 for GLTW$_{7b}$, 32 for GLTW$_{3b}$, and 64 for GLTW$_{1b}$ over WN18RR and Wikidata5M. For FB15k-237, the batch sizes are set to 32 for GLTW$_{7b}$, 64 for GLTW$_{3b}$, and 128 for GLTW$_{1b}$. Note that for all LLMs, we fine-tuned them using LoRA technology, with parameters set as follows: $r = 32$,  $dropout = 0.05$, and $target \ modules = (query, value)$.
Note that in specific experiments, the element functions $f_1$ and $f_2$ are essentially learnable embedding layers, which encode numbers into corresponding vectors. Meanwhile, the element functions of matrix $P$ and matrix $D$ are independent of each other (except for the G2G relative position).

\section{The construction strategy of $P$ and $D$ in $g$GLM}
\label{app:P_D_gglm}
\begin{figure*}[h]
  \centering
  \begin{subfigure}{1.0\linewidth}
    \centering
    \includegraphics[width=\linewidth]{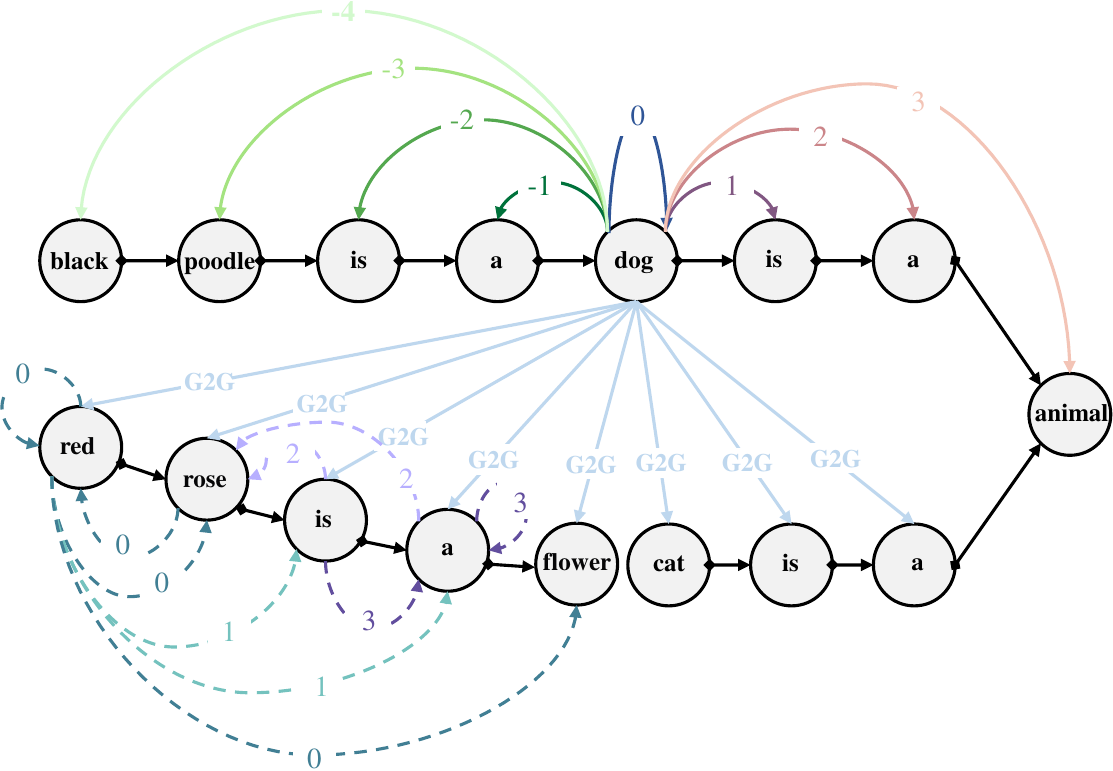}
    \caption{Levi graph of example subgraph with relative distances for \textit{dog} and distinction for <red rose, is a, flower>}
  \end{subfigure}

  \vspace{1ex} 

  \begin{subfigure}{0.48\linewidth}
    \centering
    \includegraphics[width=\linewidth]{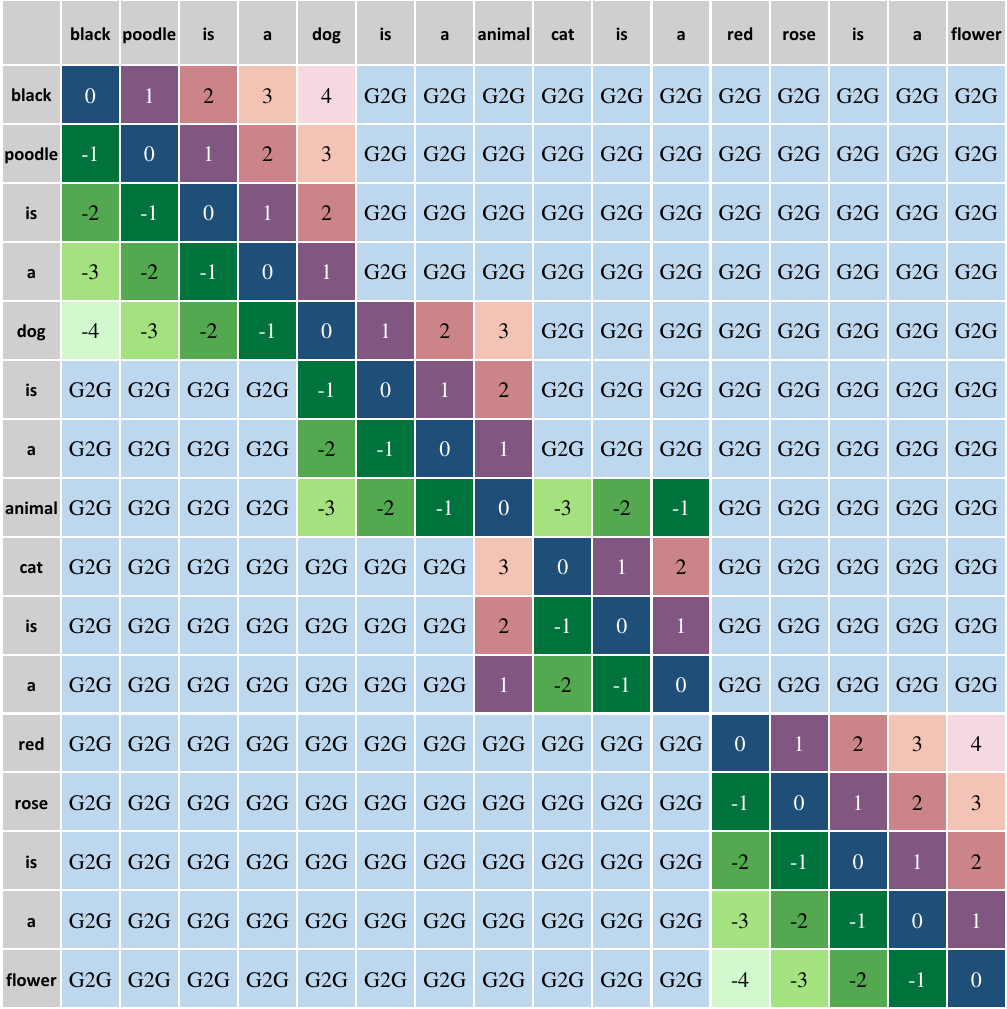}
    \caption{Relative position matrix $P$ for (a)}
  \end{subfigure}
  \hfill 
  \begin{subfigure}{0.48\linewidth}
    \centering
    \includegraphics[width=\linewidth]{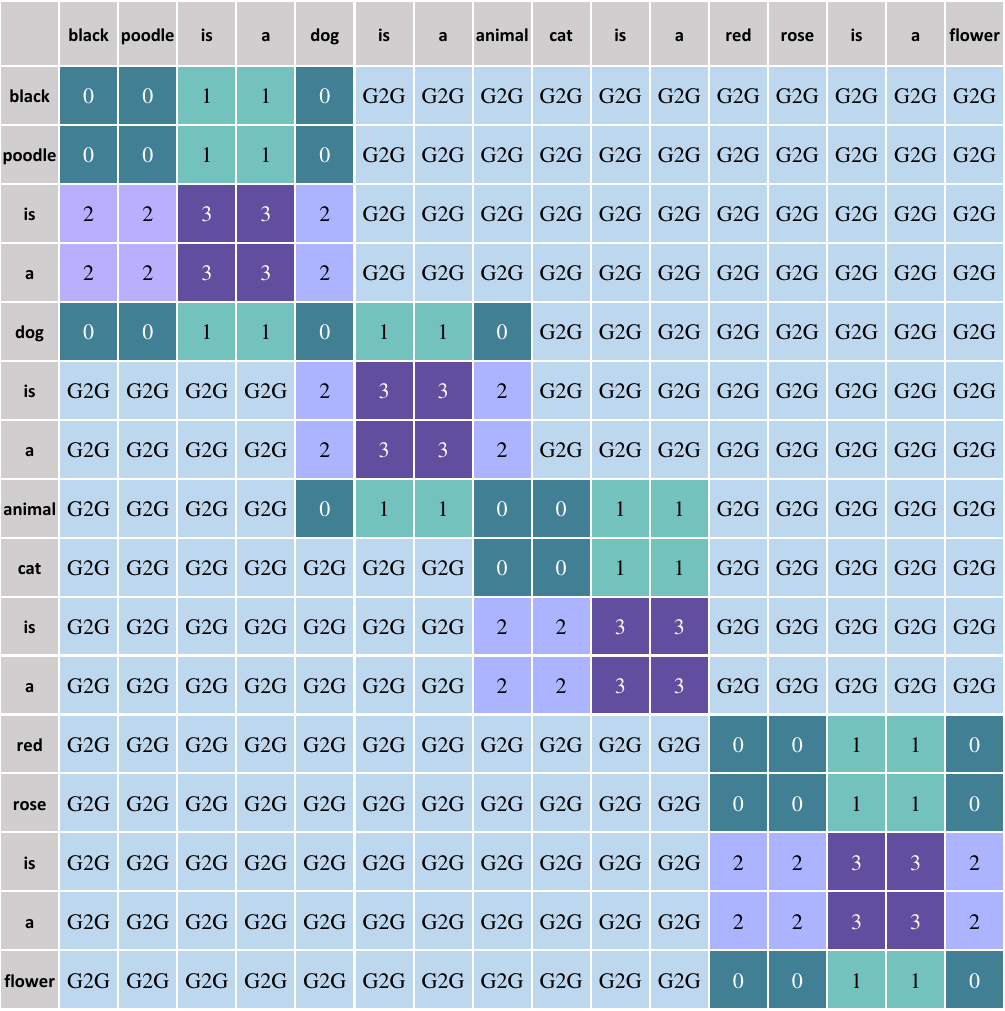}
    \caption{Relative distinction matrix $D$ for (a)}
  \end{subfigure}

  \caption{Example of subgraph preprocessing with $P$ and $D$ in $g$GLM~\cite{plenz-frank-2024-graph}. Note that entries with G2G are initialized to $+\infty$.} 
  \label{fig:p_D_gGLM}
  \vspace*{-2ex}
\end{figure*}

In this section, we introduce the positional encoding strategy of the existing method $g$GLM~\cite{plenz-frank-2024-graph}, as shown in Fig.~\ref{fig:p_D_gGLM}(a)–(b). Importantly, we integrate the proposed relative distinction matrix $D$ into $g$GLM and illustrate an example of encoding for $D$ in Fig.~\ref{fig:p_D_gGLM}(c).


\section{KG Language Prompt}
\label{app:Three_word_prompt}
We present the KG language prompt in Table~\ref{tab:three_word_prompt}. Note that the prompt in Table~\ref{tab:three_word_prompt} directly stems from MKGL, as our work is orthogonal to the design of the KG language prompt.

\begin{table*}[htbp]
  \centering
    \begin{tabular}{p{38em}}
    \toprule
    \textbf{Input:}\newline{} \#\#\# Instruction \newline{}Suppose that you are an excellent linguist studying a three-word language. Given the following dictionary:\newline{}
    Input \quad \quad \quad \quad \quad \quad \quad Type \quad \quad \quad \quad Description\newline{}
    \textit{<kgl:black poodle>} \  Head entity \quad \quad black poodle \newline{}
    \textit{<kgl:is a>} \quad \quad \quad \quad \ Relation \quad \quad \quad \quad is a \newline{}
    Please complete the last word (?) of the sentence: \textit{<kgl:black poodle>}\textit{<kgl:is a>}?\newline{}\newline{}
    \#\#\# Response: \newline{}
    \textit{<kgl:black poodle>}\textit{<kgl:is a>}
    \\
    \bottomrule
    \end{tabular}%
    \caption{\textbf{KG language prompt}: In the context of three-word Language, link prediction task corresponds to completing the sentence $hr$?. Note that we take \textit{<kgl:black poodle>} and \textit{<kgl:is a>} as a example. } 
  \label{tab:three_word_prompt}%
\end{table*}%

\end{document}